\documentclass{article}
\usepackage{authblk}
\usepackage[a4paper, margin=2.2cm]{geometry}
\usepackage{graphicx}
\usepackage[utf8]{inputenc}
\usepackage[authoryear]{natbib}
\usepackage{hyperref}
\usepackage{xcolor}
\hypersetup{colorlinks, linkcolor={black}, citecolor={gray}, urlcolor={blue!60!black}}

\usepackage{etoolbox}

\makeatletter

\pretocmd{\NAT@citex}{%
  \let\NAT@hyper@\NAT@hyper@citex
  \def\NAT@postnote{#2}%
  \setcounter{NAT@total@cites}{0}%
  \setcounter{NAT@count@cites}{0}%
  \forcsvlist{\stepcounter{NAT@total@cites}\@gobble}{#3}}{}{}
\newcounter{NAT@total@cites}
\newcounter{NAT@count@cites}
\def\NAT@postnote{}

\def\NAT@hyper@citex#1{%
  \stepcounter{NAT@count@cites}%
  \hyper@natlinkstart{\@citeb\@extra@b@citeb}#1%
  \ifnumequal{\value{NAT@count@cites}}{\value{NAT@total@cites}}
    {\ifNAT@swa\else\if*\NAT@postnote*\else%
     \NAT@cmt\NAT@postnote\global\def\NAT@postnote{}\fi\fi}{}%
  \ifNAT@swa\else\if\relax\NAT@date\relax
  \else\NAT@@close\global\let\NAT@nm\@empty\fi\fi
  \hyper@natlinkend}
\renewcommand\hyper@natlinkbreak[2]{#1}

\patchcmd{\NAT@citex}
  {\ifNAT@swa\else\if*#2*\else\NAT@cmt#2\fi
   \if\relax\NAT@date\relax\else\NAT@@close\fi\fi}{}{}{}
\patchcmd{\NAT@citex}
  {\if\relax\NAT@date\relax\NAT@def@citea\else\NAT@def@citea@close\fi}
  {\if\relax\NAT@date\relax\NAT@def@citea\else\NAT@def@citea@space\fi}{}{}

\makeatother

\usepackage{xcolor}

\title{What does it mean to represent? Mental representations as falsifiable memory patterns}

\author[1]{Eloy Parra-Barrero}
\author[2]{Yulia Sandamirskaya}
\affil[1]{Institute for Neural Computation, Faculty of Computer Science, Ruhr University Bochum, Bochum, Germany}
\affil[2]{Neuromorphic Computing Lab, Intel Labs, Intel, Munich, Germany}

\begin{document}

\maketitle

\begin{abstract}
   Representation is a key notion in neuroscience and artificial intelligence (AI). However, a longstanding philosophical debate highlights that specifying what counts as representation is trickier than it seems. With this brief opinion paper we would like to bring the philosophical problem of representation into attention and provide an implementable solution. We note that causal and teleological approaches often assumed by neuroscientists and engineers fail to provide a satisfactory account of representation. We sketch an alternative according to which representations correspond to inferred latent structures in the world, identified on the basis of conditional patterns of activation. These structures are assumed to have certain properties objectively, which allows for planning, prediction, and detection of unexpected events. We illustrate our proposal with the simulation of a simple neural network model. We believe this stronger notion of representation could inform future research in neuroscience and AI. 
\end{abstract}

\bigskip

Cognition is often described as computation over representations to yield behaviour \citep{Gallistel1990, Barack2021}. Although some might disagree with this explicit formulation, the notion of representation certainly plays a central role in the study of natural and artificial cognition. Neuroscientists invoke it when discussing almost all aspects of cognitive brain function, from sensory processing to learning and memory or decision making (e.g., \cite{Kriegeskorte2008}, \cite{Quiroga2012}, \cite{Knutson2005}). Perhaps neuroscientific research in spatial cognition highlights this point most clearly. Over the past few decades, researchers have uncovered a zoo of neuronal cell types said to represent all sorts of spatial properties: position, heading direction, speed, the presence of borders, objects, goals, etc. \citep{Hartley2014, Bicanski2020, Parra-Barrero2021}. In machine learning, researchers also concern themselves with ``representation learning” \citep{Bengio2014}. However, philosophers point out that it is not at all trivial to give a naturalistic account of representation or \textit{aboutness}. What makes my desire to drink tea be \textit{about} or directed at tea, as opposed to coffee, or to nothing at all? After all, stones and trees and hearts are not about anything, they just are what they are. As Nicholas Shea puts it, the problem of representation is to explain how states of brains or artificial systems manage to \textit{reach out} and be about anything outside of themselves \citep{Shea2018}. \\

A naive solution to this problem is to say that some cognitive system's representational vehicle, $R$, (e.g., the activation of a neuron) represents some state of the world, $S$, because $R$ is caused by $S$ and is thus indicative of it \citep{Adams2021}. However, this is clearly not enough. Smoke is often caused by, and is thus indicative of fire, but it does not “represent” fire in any meaningful way. A problem with this example seems to be that smoke does not do anything relevant within a broader system. Could we, perhaps, fix this by requiring that a representational vehicle, $R$, influences some other part of the system it belongs to in driving a response to $S$? Think of a row of falling dominoes in a Rube Goldberg machine. The fall of the second tile is indicative of the fall of the first, and triggers the fall of the third. Could we then say that the fall of the second tile ``represents'' the fall of the first tile to the third? Somehow, this example still falls short of defining a representation. To explain why the third tile falls, we only need to consider the fall of the second. This is so because whatever impact the first tile had, it is already contained in the fall of the second tile. Thus, we do not gain anything by saying that the second tile represents the first. If we want to claim that $R$ represents $S$, $R$ should not already casually imply the occurrence of $S$, because then $S$ would become superfluous in the explanation. This means that $R$ and $S$ should be causally decoupled, with $R$ being able to represent $S$ even in $S$'s absence. This causal decoupling is the key to representations and the basis for two features which are generally regarded as the hallmarks of `true' representations. The first one is the ability of representations to participate in sophisticated cognitive capacities such as remembering the past, imagining fictitious events, or predicting the future. The second one is that the use of representations is error-prone. Since $R$ represents $S$,  but is not necessarily caused by it, sometimes $R$ might \textit{misrepresent} the state of the world as being $S$ \citep{Dretske1986, Fodor1990}. Your mental representation of `snake' might be incorrectly triggered by the sight of a piece of rope. This highlights that representations are ``normative''. They are \textit{supposed} to indicate something quite specific, and they can get it right or wrong. Here is where things get complicated, because, what does it mean for a part of a system to be \textit{supposed} to indicate something? \\

A common reply to this question involves invoking some function or purpose (``telos'', in ancient Greek, from which the name of this approach, ``teleosemantics'', is derived) \citep{Neander2021, Baker2021}. $R$ is supposed to indicate $S$ because that is what enables the `proper' functioning of the system that $R$ was selected for. For instance, some fly-detecting neuron in the brain of the frog is supposed to indicate flies because that is what enables the proper feeding behavior of the frog that the neuron was selected for throughout evolution. Thus, if a frog is tricked into throwing its tongue at moving black dots on a screen, the fly-detecting neuron would have misrepresented the dots as flies, or so the story is supposed to go. \\

A problem with the teleosemantic approach is that representations tend to become defined in historical terms (e.g., the activation of fly-detecting neurons having co-occurred with flies in previous generations of frogs). This history is external to the cognitive system that contains the representation, and therefore cannot play a causal role within it \citep{Bickhard2009}. It might be useful to \textit{describe} the fly-detecting neuron as representing flies, but the representing is not doing any causal work. It would drop out of a causal explanation just like the first tile in the example of the dominoes. Furthermore, because this history is external to the cognitive system, it is only an external observer who can grasp it. So for $R$ to represent $S$, we would need another observer that can already represent $R$, $S$ and the ways in which they have been related to each other in the past. But this would lead to an infinite regress of interpretative homunculi. For these reasons, Mark Bickhard insists on the need for representations to have their representational content (what the representation is about, which defines the representation) defined internally within the cognitive system that has them \citep{Bickhard2009}. This way, the cognitive system can evaluate whether it is using a representation correctly or not, and this could account for the normativity of representations. In other words, we could say that $R$ is supposed to indicate $S$ for a system to the extent that the system itself has the ability to detect erroneous attributions of $R$. Your mental representation of snakes is supposed to indicate snakes, and not ropes, precisely because when it is caused by ropes, you can, at least in principle, later recognize the mistake.\\

This leads us to the notion that representations correspond to inferred latent structures. Locked up in the skull, the brain does not have direct access to the identity of objects and processes outside (e.g. the rope or the snake). These objects and processes are therefore `latent' or `hidden' structures (also referred to as variables, factors or causes [of the sensory input]), whose presence can only be inferred based on noisy and incomplete sensations. Note that by inference we do not mean the conscious and effortful process we engage in every other day when solving puzzles. Following Helmholtz, we conceive of inference as a pervasive and mostly unconscious process that underlies all of perception \citep{vonHelmholtz1867}. The key about inference is that the inferred latent structures are defined internally within the cognitive system, in terms of some properties that they are assumed to have (e.g., we may decide that ``snake'' refers to long, scaly animals that can bite). The latent structures are assumed to have these properties objectively, regardless of whether they are currently being accessed or not. For example, when you infer that some object lying ahead is a snake, you assume that it could move and bite you---even if it is not doing it yet. This allows you to plan actions that involve those hidden properties. For instance, you might decide to take a detour around the apparent snake. The assumption that certain properties are present also allows you to falsify your hypothesis and recognize you were mistaken. If the snake turns out to be a rope, you may realize that what you thought to be the case --that this is a scaly animal that could move and bite you-- was false. Thus, identifying representations with inferred latent structures, we can account for how representations are \textit{supposed} to indicate what they do from the perspective of the cognitive system itself, enabling representational content to play a causal role. \\

An account of representation based on inference therefore looks promising. But are there other reasons for postulating that cognitive systems infer latent structures? In principle, a cognitive system could get by without doing this. For example, the system could use a huge look-up table that tells it how to react to every possible combination of inputs. Improving on this slightly, the system could also use a very large feedforward neural network with only one hidden layer, thanks to the universal approximation theorem \citep{Hornik1989}. However, this will not work in practice because the number of possible input-output mappings is too vast. Luckily, there is a lot of structure in the world that can be leveraged to find more compact solutions. The world seems to instantiate a compositional hierarchy, in which higher-level entities or structures are composed of lower-level ones \citep{Lecun2015}. Societies contain people, which contain faces, which contain eyes, which contain pupils. Structures at all levels appear repeatedly. Societies contain people, but so do riots and ballet companies. People have faces, but so do fish. How to recognize the presence of some structure, and what that presence affords you is largely preserved across different manifestations of that structure. Thus, it would be a waste of time and resources if we had to learn how to recognize and interact with each structure independently for each type of complex it appears in. It is much more efficient to abstract away from particular occurrences and recognize the existence of this recurring latent structure. Moreover, such recurring structures, when applied across different domains could be the basis for metaphorical thinking and the extrapolation of past experiences to novel situations---arguably some of our most sophisticated cognitive abilities. \\

Interestingly, organizing latent structures in a hierarchy confers one further practical advantage. We know many different facts about each structure. We know how people look and sound like, how their knees bend, how they break your heart. Thinking about one of these facts, we can retrieve any other when relevant.  This ability requires the capacity to propagate activity from the representation of each fact to that of each other fact (guided and constrained by the current context so that we do not get flooded by all possible associations). To accomplish this without a hierarchy, if we knew $N$ facts about some structure, we would need $N^2$ (all-to-all) connections between their representations. With a hierarchy, we would just need $2N$ connections, going back and forth from the representation of the structure to those of each of its components. \\

Deep neural networks purport to exploit such a compositional hierarchy by representing increasingly more complex latent structures or factors across layers of a neural network \citep{Bengio2014}. For instance, \citep{Higgins2021} claim that a variational autoencoder fed with images of faces learns to represent latent structures such as the gender or the age of a face. So do deep neural networks infer latent structures, thus accounting for representations? We do not think so. As we saw above, what is required to talk of representations is some justification for how some state is \textit{supposed to} indicate something, even when it is not being caused by it. A possible justification involves the system assuming that certain hidden properties are present, which can be proved wrong by the system itself. However, there is none of this in a feed-forward neural network. A feed forward network is, in the end, a mathematical function that is sensitive to the presence of some pattern in the input. In the example just mentioned, the network could have learned to recognize the visual appearance of age, perhaps just wrinkliness and longer ear lobes. This wrinkliness is nothing hidden that the network could be mistaken about, it is either there, ``in plain sight'', or it is not. Thus, if the network, for example, classifies a young face with painted wrinkles as ``old'', the network did not make any mistake from its own perspective. The network just correctly indicated the presence of wrinkles.\\

What about approaches that focus on Bayesian inference? In these approaches, the brain is purported to (or artificial systems are constructed such that they) learn probabilistic generative models of the environment. These models are then used to infer the most probable causes of the inputs \citep{Fiser2010, Tenenbaum2011}. Often, the inferred causes will correspond to latent structures whose identity the cognitive system could be mistaken about. But this is not necessarily so. Consider a toy world where all an agent can do is look at one of two balls of different colors (a red ball and a green ball), and then `infer' which of the two it is looking at. Even though this is a trivial task, nothing prevents one from applying Bayesian inference to solve it. However, in this kind of situation, the inferred causes (e.g., ball \textit{A} and ball \textit{B}) are not much more than labels for visible properties (e.g., redness and greenness). As in the case of the network detecting wrinkles, the inferred causes here do not refer to anything with hidden properties that the agent could possibly be mistaken about. Thus, Bayesian inference does not seem to get to the crux of what is required to infer latent structures and form proper representations. \\

\begin{figure}[htb]
    \centering
    \includegraphics[width=\textwidth]{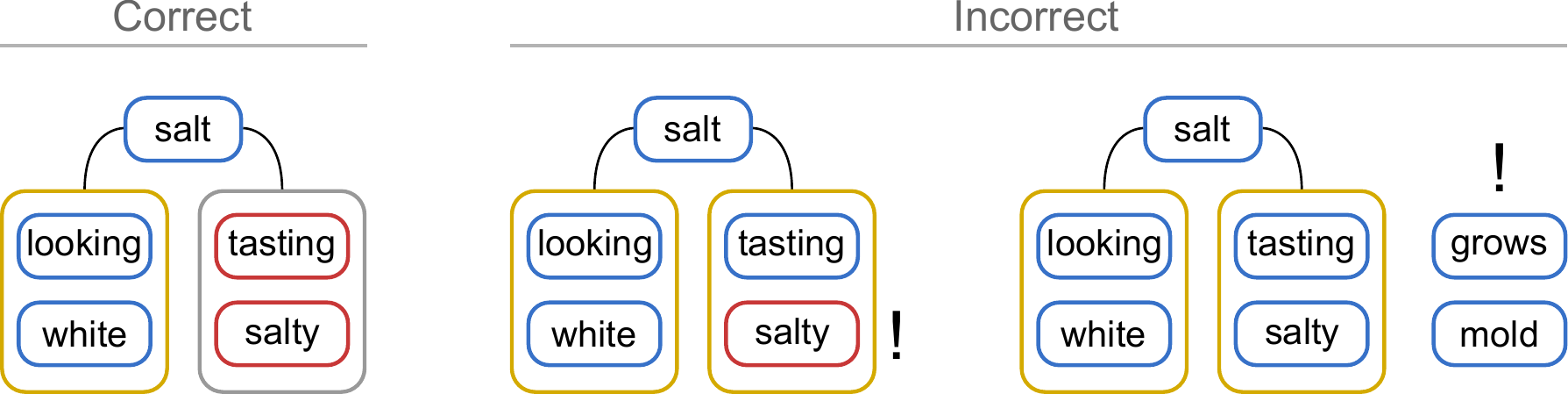}
    \caption{\textbf{Schematic representation of the concept of salt.} A node that represents a white salty substance (``salt'') is defined by two conditional bistable patterns: \{`tasting', `salty'\} and \{`looking', `white'\}. Each element can be present (blue) or absent (red). If half of the elements in the pattern are present, the pattern becomes applicable (gold). It is correct to infer that ``salt'' is present when at least one of the patterns is applicable and complete, no pattern is applicable but incomplete, and no unexpected elements appear.}
    \label{salt}
\end{figure}

To develop an approach to representation that has the required properties, we must reflect on what it means for something to be a latent structure. Something is latent because it does not manifest itself completely, at least not all of the time. Take the example of salt. Its various aspects reveal themselves under different circumstances. When looking at it, you see something white. When putting it into your mouth, you taste saltiness. When eating too much, you get thirsty. When adding it to meat, it helps to preserve it. These are different patterns of interaction between the latent structure that we call salt and ourselves or other latent structures. Each pattern of interaction is composed of certain elements. For example, \{`tasting', `salty'\} is one pattern, and \{`ate', `too much', `thirsty'\} is another. When a certain pattern of interaction is taking place, as recognized by the fact that a certain fraction of its components are present, the whole pattern is expected to be there. Thus, under the hypothesis that something is salt, if `tasting' is present, so should `salty', and vice-versa; if \{`ate', `too much'\} is present, so should `thirsty', or if \{`ate', `thirsty'\} is present, so should `too much'. Perhaps we can call patterns like \{`ate', `too much', `thirsty'\} \textit{conditional bistable patterns}. The patterns are conditional on the assumption that salt is involved in explaining the situation. Based on this condition, the elements of the pattern should either be mostly absent (the pattern is `OFF') if the interaction is not taking place, or mostly present (the pattern is `ON') if the interaction is taking place, but not somewhere in between. Thus the patterns are bistable. Latent structures seem to be defined by sets of such conditional bistable patterns. We could then conceive of cognitive systems that explicitly encode and store these patterns in memory, and use them to infer latent structures. A good heuristic for the inference process could be that a latent structure is recognized to be present if at least one of its conditional bistable patterns is ON, and none of them is in an unstable state.  For example, you may infer that something is salt based only on looking at some white grains.  If you then taste them and they turn out to not to be salty, the interaction pattern characterized by \{‘tasting’, ‘salty’\} is recognized as being applicable (half of elements are present) but failing to obtain (‘salty’ is missing), and you should discard the hypothesis that what you had in front of you was salt. Missing elements are not the only kind of negative evidence for your hypothesis. The presence of unexpected elements is equally revealing. Salt should not combust, or grow mold, or make people turn blue. If any of these things happen, you will also detect a violation and realize that you were not dealing with salt. However, there are infinitely many things that should not happen in any given context, so it is impossible to explicitly learn all of them. Thus, observing anything that deviates from the relatively small set of known conditional bistable patterns should by default count as evidence against something being what you thought it was. These combinations of conditions for the (simplified) example of representing salt are summarized in Fig.~\ref{salt}. \\

The unexpected sensations that count as evidence against an inferred latent structure are those that are not accounted for or “explained away” by any of the currently inferred latent structures. A natural way of detecting them is using something akin to predictive coding \citep{Huang2011}, where error units indicate the difference between the inputs and top-down predictions based on inferred latent structures. In our framework, such error units should be of two kinds to detect the two kinds of violations described above: either some element should be present but is not, or it is present but it should not be.  \\

\begin{figure}[htb!]
    \centering
    \includegraphics[width=\textwidth]{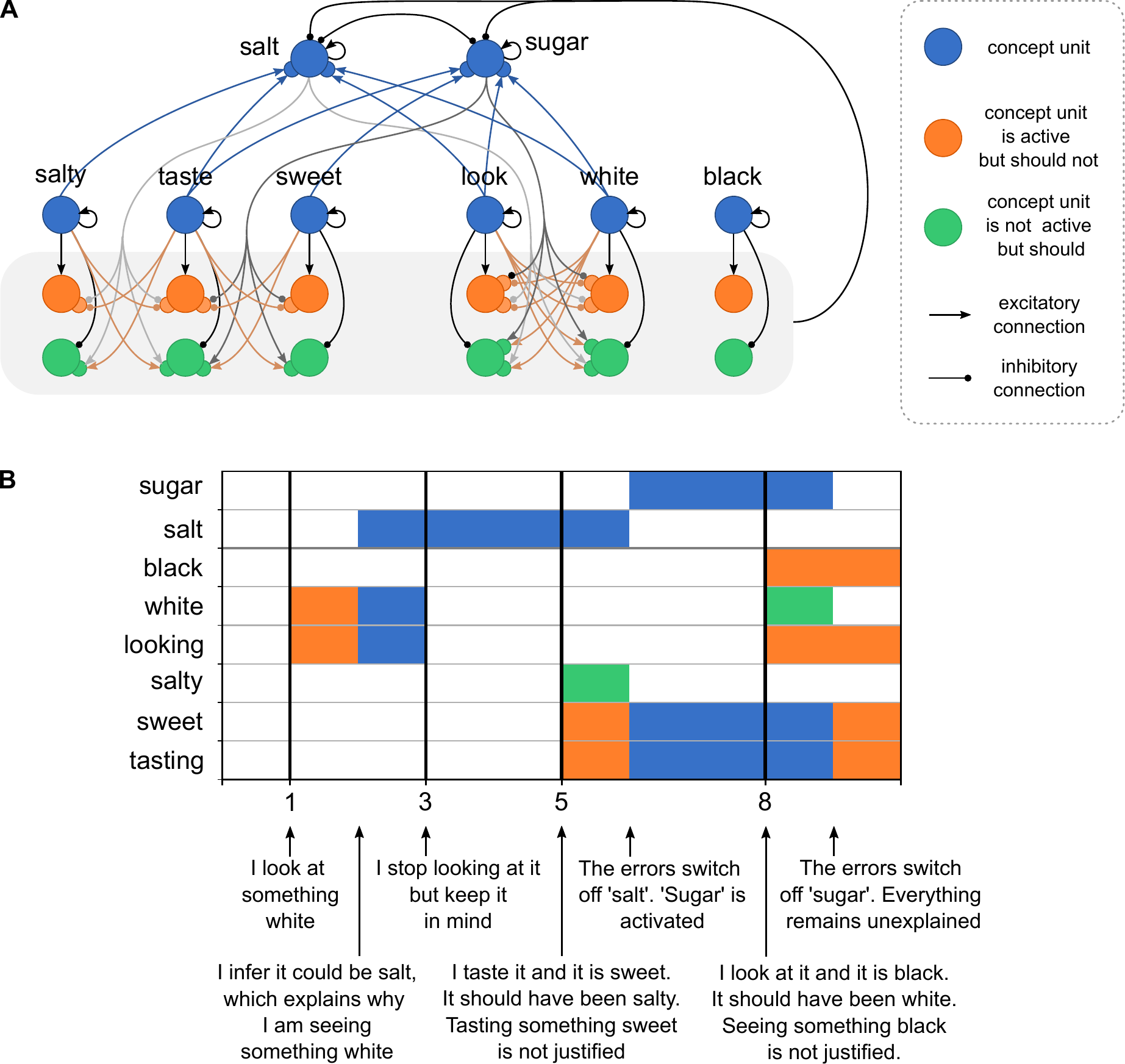}
    \caption{\textbf{A neural circuit implementation of representations as sets of conditional bistable patterns.} 
    \textbf{A}: The neural network implements the computations required for the inference of ``salt'' and ``sugar''. See main text for details. 
    \textbf{B}: A simulation of the network in action. Blue, orange and green correspond to the activation of the concept and error units, respectively, following the same color code as above. Note, however, that when the orange unit is active, so is the blue one `underneath'. The code for the simulation can be found at \url{https://github.com/EloyPB/concepts}. 
    }
    \label{circuit}
\end{figure}

To make our proposal more concrete, in Fig.~\ref{circuit} we sketch a neural circuit that fleshes out the ideas put forth above, relying on the two types of error units. The neural circuit shown in the figure implements the computations required for the inference of the concepts ``salt'' and ``sugar''. The concept units representing these two inferred latent structures (blue circles in the figure) can self-sustain their activation through recurrent self-connections and compete with each other for explaining the inputs via lateral inhibition. ``Salt'' becomes active when either ``salty'' and ``tasting'' and/or ``looking'' and ``white'' are active. These conjunctions are detected by nonlinearities in dendritic compartments (depicted as small colored circles). This mechanism is inspired by the finding that dendrites can implement nonlinear functions, making individual neurons much more complex computing structures than conventionally assumed in artificial neural networks \citep{Beniaguev2021}. The concept units at the bottom level of the small hierarchy in our example (``salty", ``taste", etc.) have corresponding error units that signal whether the concept is active but should not (orange circles), or whether it is not active but should be (green circles). These units also have nonlinear dendritic compartments which only become active when both of their inputs are active. When the patterns are in a stable state (for example, when ``salt'', ``salty'' and ``tasting''  are all active, or when ``salt'' is active but both ``salty'' and ``tasting'' are not), the excitation and inhibition arriving at the error units cancel out and the error units remain silent. Otherwise error units become active, indicating what is missing or what should not be present. The sum of all error units inhibits the concept units at the top level. One can imagine this network motif being repeated across multiple layers. The ``salt'' and ``sugar'' units could have error units themselves, which would receive top-down input from a layer higher in the hierarchy that contains, for instance units representing ``anchovies'' or ``cake''. The lower part of Fig.~\ref{circuit} shows the activation of the network structures when representing different perceived situations. \\

The presented neural circuit is just an exemplary one. We provide this suggestion here to facilitate the understanding of our proposal with a concrete example, and to stimulate further thought about possible implementations. Regardless of the details, a neural circuit that relies on calculating different types of errors between observed and expected phenomena brings important advantages for cognitive and machine learning architectures, apart from making proper representations possible. For one, checking for things out of the ordinary is arguably one of the hallmarks of common sense---something that current standard approaches in AI are notoriously bad at. Second, the error signals are useful for guiding attention and learning. And third, the error signals could be used to drive behavior. For instance, if you are trying to cure some meat, one of salt's error units could light up, signalling that salt is required and missing so that you can go and fetch some.\\ 

To conclude, we have proposed that mental representations can be decoupled from the input and can misrepresent (from the point of view of the system itself) because they are the result of a fallible process of inferring latent structures. Inferring these latent structures is necessary in practice, as it leads to a much more compact and flexible cognitive system that can plan actions based on hidden properties of the inferred latent structures. Deep learning and Bayesian approaches both provide valuable components for the construction of such a cognitive system. The former offers effective ways of learning hierarchies of increasingly more complex structures, whereas the latter offers sophisticated probabilistic reasoning capabilities. However, we argue that additional mechanisms are required to ensure that cognitive systems truly infer latent structures they could be mistaken about. These mechanisms could build on our proposal that latent structures are defined in terms of sets of conditional bistable patterns. Explicit knowledge of these patterns could be used in the inference process, where the detection of unexpected elements would also play a fundamental role. We hope these views will help move the field of AI over to systems that build `true' representations, which would lead to more intelligent systems capable of reasoning, abstraction and autonomous learning. From the neuroscience perspective, we should be more careful not to call every signal that correlates with the input a representation. A tuning curve or a receptive field is not enough. We should therefore design experiments that test whether representations can be decoupled (for example, for their use in planning or imagination), and whether they can elicit error signals. The proposed computational structure could also help explain and be validated by the connectivity patterns across cortical layers of the neocortex that form what has been dubbed a ``canonical microcircuit'' \citep{Douglas1989}. The wiring of neurons across layers of the neocortex seems to perform some kind of predictive-coding computation with top-down, bottom-up, and lateral connections and signal flow \citep{Bastos2012}, which might be the biological neural substrate behind the ability of the brain to represent items in the external world in the deep and fundamental way we have argued for in this paper.

\subsection*{Acknowledgements}
Many of the ideas expressed here originated in debates with José R. Donoso around the notion of mental representation, and were sharpened by further discussions with Alexander M. Hölken, Daniel Sabinasz, Xiangshuai Zeng, Nicolas Diekmann, Laia Serratosa, Olya Hakobyan and Sen Cheng among others. We thank them for their insightful engagements and friendly pushback. 

\bibliography{mental}
\bibliographystyle{apalike}

\end{document}